\pdfoutput=1

\documentclass[11pt]{article}

\usepackage{ACL2023}

\usepackage{times}
\usepackage{latexsym}
\usepackage{amsmath}
\usepackage{amssymb}

\usepackage[T1]{fontenc}

\usepackage[utf8]{inputenc}

\usepackage{microtype}

\usepackage{inconsolata}

\title{Prompt Generate Train (PGT): Few-shot Domain Adaption of Retrieval Augmented Generation Models for Open Book Question-Answering}

\author{C. S.  Krishna \\
Microsoft \\
\texttt{krishnacs@microsoft.com} 
}

\begin{document}
\maketitle

\begin{abstract}
We propose a framework – Prompt, Generate, Train (PGT) – to efficiently develop a generative question-answering model for open-book question-answering over a proprietary collection of text documents. The framework adapts a retriever augmented generation (RAG) model to the target domain using supervised fine-tuning and reinforcement learning with synthetic feedback in a few-shot setting. This, we hypothesize, will yield an aligned, uncertainty calibrated model that is competitive with GPT-4 based in-context retrieval augmented generation in generating relevant answers at lower serving costs. 
\par The framework’s synthetic generation pipeline will generate synthetic training data comprising <passage, question, answer> tuples using an open-source LLM and a novel consistency filtering scheme. The pipeline will be designed to generate both abstractive and extractive questions that span the entire corpus. The framework proposes to fine-tune a smaller RAG model comprising a dense retriever (ColBERTv2) and a smaller sized LLM on the synthetic data-set. In parallel, the framework will train a Reward model to score domain grounded answers higher than hallucinated answers using an \textit{a priori} relevance ordering of synthetically assembled samples. In the next phase, the framework will align the RAG model with the target domain using reinforcement learning (Proximal Policy Optimization). This step may improve the RAG model’s ability to generate grounded answers and ignore out of domain questions. In the final phase, the framework will calibrate the model’s uncertainty for extractive question-answers.
\end{abstract}

\section{Introduction}
A common use-case within enterprise settings is to expose a question-answering service over a proprietary corpus comprising a collection of text documents. The application, in response to the client’s question, must be able to generate a factually grounded, attributed response by condensing information from a set of relevant documents in the underlying corpus. If the answer cannot be synthesized from these documents, the system should respond with a “cannot answer” rather than generate a misleading or factually incorrect response (referred to as hallucination in the literature). An additional requirement is to economize development and serving costs by using smaller LLM architectures (<10 Bn parameters).

\par It has been shown that specialized task and target domain data specific pre-training and/or fine-tuning enables smaller sized LLMs to outperform in-context learning with LSLMs \cite{izacard2022few,hsieh2023distilling}. However, smaller LLMs have inherent weaknesses. \cite{gudibande2023false} demonstrated that certain properties such as chain of thought reasoning \cite{wei2022chain} only emerge at higher scales. A smaller LLM will therefore struggle to outperform LSLMs such as GPT-4 on questions that require reasoning ability. To address this limitation, the PGT framework needs to include a procedure for model uncertainty calibration. Informally speaking, the model must “know when it knows the answer” and “know when it doesn’t know the answer”. We make the notion of “knowing when the model knows” and “knowing when it doesn’t know” more precise in the section on uncertainty calibration. This is a desirable feature since it enables easier integration into a cascading systems such as FrugalGPT \cite{chen2023frugalgpt} where the RAG model’s answer can be surfaced only when the model is confident of its answer, else the client's question can be passed on to a human or alternate model.

\par Further, smaller LLMs are more prone to hallucination [ref?]. The framework design, therefore, needs to address hallucination mitigation. Given this perspective, we list the following design goals for the PGT framework:
\subsection{Design Goals}
\begin{itemize}
\item	\textbf{Few-shot Adaption}: The framework needs to adapt the RAG model to the target domain in a few-shot setting. i.e., it must do so without access to a large volume of manually annotated <question, answer> tuples on which the model can be fine-tuned.
\item	\textbf{Serving Cost Economization}: PGT models must be cheaper to serve and more accurate than systems based on GPT-3.5/4 powered in-context retrieval augmented generation. Since serving cost is a function of model size, we choose models with less than 10 Bn parameters.
\item	\textbf{Hallucination Mitigation}: The framework should yield a model that only generates answers based on the underlying corpus. For out of domain questions or questions which cannot be answered based on the underlying corpus, the model should be able to generate “cannot answer this question based on given information” rather than hallucinate a baseless answer.
\item \textbf{Uncertainty Calibration}: The framework should yield models calibrated for uncertainty, at least for a certain class of questions, such as extractive or yes/no questions. The application layer can then rely on the model’s confidence in its generated answer to decide whether to surface answer to the end-user or take a pass on the question.
\end{itemize}

\section{Related Work}
The dominant framework to develop models for open-book question answering is based on in-context retrieval augmented generation \cite{ram2023context}(Bing Search): in response to the user question, a dense or a sparse retriever fetches relevant documents. A prompt is populated by the user’s question and the set of fetched documents. This prompt is then fed to an LSLM such as GPT-4 to generate the response. This framework does not require domain adaptation since the LSLM has been extensively pre-trained on public domain corpora such as Wikipedia, Common-Crawl etc. Forcing the LLM to rely on the fetched documents also constrains the LLM to generate grounded answers to a certain extent, although LLMs such as GPT-4 can still be factually wrong in subtle ways [refs?].
\par While such a framework is suitable for open-domain question services such as search engines, where the service is expected to answer questions from any domain, we argue that it is overkill for closed-domain question answering. In the latter setting, the model is only expected to generate answers for in-domain questions. Hence, it stands to reason that the LLM’s knowledge base merely needs to cover the target domain for which a smaller LLM can suffice, motivating the feasibility of adapting a smaller LLM for the target domain. 
\par A popular approach for domain adaptation for question-answering is to train a RAG architecture \cite{pmlr-v119-guu20a, lewis2020retrieval} comprising a retriever and generator on annotated training data in the form of <question, answer> tuples from the target domain. \cite{shi2023replug}, on the other hand, only trains the retriever component of the RAG architecture. \cite{izacard2022few} showed that jointly fine-tuning the RAG model on as few as 1024 samples from the target domain improves performance relative to GPT-3 based in-context retrieval augmented generation. They also empirically demonstrated that performance increases when the model is fine-tuned on larger sized data-sets.
\par In most real world settings, such manually crafted data will not be available in sufficient volume or quality. An emerging body of work has adapted dense neural retrievers to the target domain by training on synthetically generated training data comprising <query, relevant passage> pairs \cite{vespa23ranking, dai2022promptagator, saad2023udapdr, abonizio2023inpars}. We extend this line of work to domain adaptation of full-fledged RAG models in a few-shot/zero-shot setting.

\section{Methodology}
The PGT framework utilizes two LSLMs – GPT-4 and Flan-TF XXL – to generate question answer pairs over all documents in the corpus. GPT-4 is used to generate a seed set of Y samples for extractive and abstractive question-answering formats. The Flan-TF XXL instance leverages the seed set to generate more samples. This design economizes the cost of generating the training dataset. The RAG model is then trained on this dataset, using both supervised fine-tuning and reinforcement learning. 
\par To generate the synthetic data-set with the LSLMs, we split documents into document segments such that the document segment size is bounded by the permissible context window size for the LSLM (1024-2048 tokens). We create an index $I_{1}$ over these document segments to facilitate random retrieval of a segment. Similarly, we maintain another index $I_{2}$ over document segments where the document segment size is upper bounded by the context window size of the RAG model’s generator.

\subsection{PGT Steps}
The PGT framework consists of the following steps:
\begin{enumerate}
    \item Phase 1a (optional): Adapt the retriever component of the RAG model to the domain by training on the Inverse Cloze Task \cite{khattab-etal-2021-relevance, lee2019latent}.
    \item Phase 1b: Synthetic Training Data Generation pipeline: In this phase, we generate two sets of data to fine-tune the RAG model and the Reward model.
    \item Phase 2: This comprises 2 training procedures that can be executed in parallel.
    \begin{enumerate}
        \item RAG model supervised fine-tuning 
        \item Reward model training
    \end{enumerate}
    \item Phase 3: Use of reinforcement learning (PPO) to align the RAG model to the target domain, with the Reward model generating a reward score for relevance.
    \item Further fine-tuning of the RAG model for uncertainty calibration for certain classes of questions.
\end{enumerate}
We can repeat steps 2, 3 and 4 for a few iterations.
\section{PGT Components}
\subsection{RAG-Retriever}
We choose ColBERTv2 \cite{khattab2020colbert} as the dense retriever in the RAG model. Dense retrievers work best with smaller sized documents of not more than $400$ tokens. On the other hand, feeding the generator with chunks rather than whole documents can lead to context fragmentation and degrade the generated answer’s quality. To resolve this trade-off, we modify the retrieval procedure as follows: we maintain a third index $I_{3}$ in which each document segment from $I_{2}$ is split into smaller sized, mostly disjoint chunks of token size $\sim 300$ tokens. The retriever component of the RAG model indexes into $I_{3}$  to fetch relevant chunks. 
\par For a given chunk $c$ of a document and the question $q$ we can compute the similarity score $S(q,c)$ using the  ColBERTv2 model instance. Let a document segment $d \in I_{2}$ correspond to chunks $(c_{1},..,c_{n})$ indexed in $I_{3}$.
We define the probability of fetching $d$ given a question $q$ to be proportional to the maximum similarity score over all the document chunks:
\begin{equation}
P_{\eta}(d|q)  \propto exp(max(S(q,c_{1}), ... , S(q,c_{n})))
\label{eqn:retrieverprob}
\end{equation}
Here, $\eta$ refers to the tunable parameters of the retriever. This design and retrieval mechanism, we hypothesize a) ensures tighter coupling between the retriever and the generator, and  b) mitigates context fragmentation, thereby improving retrieval and generation quality. We can optionally add a lexical similarity signal based on BM25 \cite{ma2020zero} to the similarity score.
\subsection{RAG Generator}
We choose a pre-trained instance of the Flan-T5 encoder-decoder architecture for answer generation. The encoder encodes the question and passages fetched by the retriever. The decoder, conditioned on the encoding, autoregressively computes the likelihood of generating the answer as
\begin{equation}
P_{\phi}(a|q, d)  = \prod_{i} P_{\phi}(a_{i}|q, d, a_{ < i})
\label{eqn:retrieverprob}
\end{equation}
\subsection{Reward model}
We also initialize a Reward model using a pre-trained BERT instance \cite{devlin2018bert}. The Reward model is trained to generate a relevance score, given the passage, question, and answer. 

\section{Synthetic Data Generation}
\subsection{Seeding with GPT-4}
In this phase, we prompt GPT-4 to generate two sets of  Y <passage, question, answer> tuples across the following question formats: extractive (EX) and abstractive (AB) \cite{khashabi-etal-2020-unifiedqa}.
\par We randomly sample document segments using $I_{1}$. We try to preserve documents boundaries as much as possible so that the LLM has a coherent passage as the basis for generating a question-answer pair. We append a q-a format specific prompts (Appendix-I) for each of the q-a formats. This prompt along with the passage is fed to GPT-4 to generate the corresponding answer. We use the seeding set of $Y$ <passage, question, answer> pairs in turn to prompt Flan-TF XXL to generate $Z$ question answer pairs across both formats.

\subsection{Generate non-matching question answer pairs}
 It is important to train the model to generate a “can’t answer based on given references” response as well if the fetched passages cannot generate the required answer. To this end, we also fetch the top $K’$ matching chunks using the retriever, remove the chunks that were used to generate the answer, and assemble a non-matching passage, $p'_{c}$.
We then prompt the LSLM to generate a rationale $r_{c}$ for why the given question $q_{c}$ cannot be answered given context $p'_{c}$:

\[ a'_{c}  \leftarrow LSLM(p'_{c}, q_{c}) \]

We assemble a non-matching question-answer pair for every matching question answer pair $(p'_{c}, q_{c}, a'_{c})$ which we refer to as a non-matching tuple. During training, the ratio of number of non-matching to matching question-answer pairs is a design parameter that can be set based on downstream requirements – intuitively, if we want the model to err on the side of caution, we should include as many non-matching question answer pairs as matching question answer pairs.

\subsection{Generation with Flan-T5 XXL}
In this phase, we tap Flan-T5 XXL to generate more training data, using the seed set from the previous phase to sample demonstration exemplars.
\begin{enumerate}
\item Start with the synthetic training data-set $T$ populated via GPT-4.
\item Concatenate a prefix prompt $P$ consisting of $N$ passage-question-answer tuples sampled without replacement from $T$ along with the meta prompt:

\[ P  = [(p_{1}, q_{1}, a_{1})...(p_{N}, q_{N}, a_{N})] \]

\item 	Select a candidate passage $p_{c}$ at random (alternatively, based on ideas presented in \cite{liu-etal-2022-makes} from $I_{1}$ and prompt Flan-T5 XXL to generate a candidate question-answer pair:

\[ (q_{c}, a_{c}) = FlanT5 XXL([P; p_{c}]) \]
\end{enumerate}
\subsection{Consistency Filtering}
We add $(p_{c},q_{c},a_{c})$ to $T$ if it meets the following consistency filtering criteria:
\begin{enumerate}
    \item 	For the generated question $q_{c}$ the domain adapted retriever’s top $K$ fetched documents should span the passage $p_{c}$. If not, drop this tuple else proceed to the next step.
    \item 	We again prompt Flan-T5 XXL/GPT-4 to generate an answer $a'_{c}$ to the question $q_{c}$ based on the top $K$ fetched chunks from the previous step. Retain this sample only if there is a high semantic overlap between $a_{c}$ and $a'_{c}$
    \item Confidence based threshold \cite{abonizio2023inpars}: the normalized log-probability of generating a question-answer pair for the candidate passage exceeds a threshold.
    \item Uncertainty based threshold (optional): Use a suitable measure of the uncertainty \cite{lin2023generating} of the generated sample. Accept the sample only if the uncertainty measure is less than a threshold.
    \item If the sample passes the consistency filtering steps, add it to $T$
  
\end{enumerate}
\section{Phase 2a: RAG model Supervised Finetuning}
We present a new log-likelihood function, \emph{in-context RAG-token model likelihood}, that combines RAG-token model likelihood \cite{lewis2020retrieval} with in-context RALM learning \cite{ram2023context}:
\begin{equation}
\begin{split}
\begin{aligned}
P_{\eta, \phi} \left(a|q; K, L, S \right ) = \\
\prod_{i=0}^{n_{S}-1}\prod_{j=1}^{S} \sum_{k \in top K}
P_{\eta} \left ( d_{i, k} | [q; a_{Si}^{L}]\right ) \\
P_{\phi} \left ( a_{Si+j}|[q;d_{i,k}; a_{<(Si+j)}] \right )
\end{aligned}
\end{split}
\label{eqn:incontextRAMtoken}
\end{equation}
The design parameter $S$ is the stride size which determines after how many steps we refresh the context by fetching relevant documents conditioned on the question and a subset of the answer prefix. The design parameter $L$ decides how many of the most recently generated tokens in the prefix to consider for context augmentation. $a_{Si}^{L}$  refers to the $L$ most recent tokens in the answer from the $Si$'th position and going backwards. $K$ decides how many documents to marginalize over in generating the next token at each step. $d_{(i,k)}$ refers to the $k$’th document-segment fetched by the retriever via $I_{2}$ in the $i$’th stride. $n_{S} =  n⁄S $ determines the number of fetches during retrieval, where $n$ is the token length of the answer. Note that when $S = 1$, $L = 0$, this function degenerates into the RAG-token model likelihood function. 
\par We intuit that the in-context RAG-token model likelihood provides the flexibility for the retriever to fetch the right document segments conditioned on the evolving answer. This in turn improves the quality of the conditioned generation. By marginalizing over top K documents during training at every generation step, we improve the ability of the retriever to discern relevant from irrelevant documents.

\subsection{Generation Procedure}
The transition probability associated with generating the next token in the answer is given by:
\begin{equation}
\begin{split}
\begin{aligned}
P_{\eta, \phi} \left(a_{Si+j}]|[q; a_{<(Si+j)}; K, L, S \right ) = \sum_{k \in top K}  \\
P_{\eta} \left ( d_{i, k} | [q; a_{Si}^{L}]\right ) P_{\phi} \left ( a_{Si+j}|[q;d_{i,k}; a_{<(Si+j)}] \right )
\end{aligned}
\end{split}
\label{incontextRAMtokenGenerator}
\end{equation}
This can be plugged into a standard beam decoder to generate the answer. 

\section{RAG Alignment Training}
Reinforcement learning with human feedback (RLHF) training \cite{NEURIPS2022_b1efde53, kadavath2022language}  further adjusts model parameters so as to generate answers that are aligned to human preferences. However, this procedure requires human preference feedback on answers, which may not always be available. Our goal is limited to aligning the RAG model such that its answers are grounded in the underlying corpus. Towards this end, we adapt RLHF but without recourse to human preference feedback to design a new technique – \emph{Reinforcement Learning with Synthetic Feedback} (RLSF).
\subsection{Phase 2b: Reward model Training}
Let $S \left ( \left ( p_{c},q_{c},a_{c} \right ) \right ) \in \mathbb{R}$  be a measure of the relevance of the response, given the context and the question. The relevance score should be low if the answer is hallucinated, factually incorrect or not grounded in the underlying context, and high otherwise. We want to train a Reward model that can estimate the relevance of a model’s generated response, given the context and question.
\par We set up the training data-set for Reward model training as follows. For every pair of matching and non-matching tuples, $(p_{c},q_{c},a_{c})$, $(p'_{c},q_{c},a'_{c})$, we can assemble additional tuples,$(p'_{c},q_{c},a_{c})$, $(p_{c},q_{c},a'_{c})$ to assemble a composite tuple:

\[ \tau  = \left ( \left  ( p_{c},q_{c},a_{c} \right )  \left  ( p'_{c},q_{c},a'_{c} \right ) \left  ( p'_{c},q_{c},a_{c} \right )  \left  ( p_{c},q_{c},a'_{c} \right ) \right ) \]

 We assemble an alternate data-set of such tuples $T^{*} = \{ \tau^{1},...,\tau^{N} \} $ for training the Reward model. For a given $\tau$, we fix the following orderings based on relevance of the answer, given the question and context:
 \begin{equation}
\begin{split}
\begin{aligned}
S \left ( \left (p_{c},q_{c},a_{c} \right ) \right ) > S \left ( \left ( p'_{c},q_{c},a_{c} \right ) \right )    \\
S \left ( \left (p_{c},q_{c},a_{c} \right ) \right ) > S \left ( \left (p'_{c},q_{c},a'_{c} \right ) \right )  \\
S \left ( \left (p'_{c},q_{c},a'_{c} \right ) \right ) > S \left ( \left (p'_{c},q_{c},a_{c} \right ) \right )  \\
S \left ( \left (p'_{c},q_{c},a'_{c} \right ) \right ) > S \left ( \left (p_{c},q_{c},a_{c} \right ) \right )  
\end{aligned}
\end{split}
\label{eqn:relavenceorderings}
\end{equation}

We train the Reward model by minimizing the following contrastive loss function:
 \begin{equation}
\begin{split}
\begin{aligned}
loss(\theta) = 
-\frac{1}{4} \mathbb{E}_{\tau \sim T^{*}} [ \log \left ( \sigma \left ( RM_{\theta} \left ( \tau_{1} \right ) -  RM_{\theta} \left ( \tau_{3} \right ) \right ) 
\right )  \\
+ \log \left ( \sigma \left ( RM_{\theta} \left ( \tau_{1} \right ) -  RM_{\theta} \left ( \tau_{4} \right ) \right ) 
\right )  \\
+ \log \left ( \sigma \left ( RM_{\theta} \left ( \tau_{2} \right ) -  RM_{\theta} \left ( \tau_{3} \right ) \right ) 
\right )  \\
+ \log \left ( \sigma \left ( RM_{\theta} \left ( \tau_{2} \right ) -  RM_{\theta} \left ( \tau_{1} \right ) \right ) 
\right )  ] \\
\end{aligned}
\end{split}
\label{eqn:rewardmodelcontrastiveloss}
\end{equation}

\subsection{Phase 3: Alignment using Reinforcement Learning (PPO)}
We use proximal policy optimization (PPO) \cite{schulman2017proximal} to further finetune the student LLM with the Reward model providing the reward signal for relevance of the answer as follows: We sample a passage-question pair using the question-generation pipeline from Phase I or optionally sample a tuple from $T$ to yield a passage-question tuple $(p,q)$. The RAG model then generates the answer conditioned on the passage, using Equation \ref{incontextRAMtokenGenerator} with $K = 1$. The Reward model scores the answer for relevance. We then finetune the RAG model w.r.t parameters $\phi$ of the generator by minimizing the PPO objective:
\begin{equation}
\begin{split}
\begin{aligned}
loss(\phi) = RM_{\theta}((p,q, a))  \\
- \beta \log \left ( \frac{P_{\eta, \phi}(a|q, p)}{P_{\eta, \phi^{'}}(a|q, p)} \right )
\label{eqn:ppo}
\end{aligned}
\end{split}
\end{equation}

\section{Phase 4: Uncertainty Calibration}
We want the student LLM to be calibrated for uncertainty. Informally, the model should “know when it knows the answer” and “know when it doesn’t know the answer”. We make this precise based on the definition of calibration outlined in \cite{guo2017calibration}. Let $p_{M} (a|q,p)$ be the probability assigned by the model that the answer it generated given the question and context is the correct response. Then, the model is perfectly calibrated if:

\[ P(a|p_{m} = p) = p, \forall p \in [0, 1] \]

To calibrate the model, we train the model for predicting whether the answer it generated given the question and supporting evidence is correct or wrong. We do so by maximizing the “indirect logit” \cite{lin2022teaching}, the log-probability associated with the model predicting “correct” or “wrong” for it’s own answer, given the question and supporting evidence from the corpus. 
\par We present the recipe below for uncertainty calibration: We use the RAG model to generate an answer to the question using a beam generator and transition probabilities using Equation \ref{incontextRAMtokenGenerator}:
\[ a \leftarrow P_{\eta, \phi}(a | a; K, L, S)\]
We then use the retriever component of the RAG model to fetch the top $M ( \sim 3)$ document segments that were used to generate the answer:
\[ (d_{1},..., d_{M}) \leftarrow P_{\eta}(. | [q;a])\]

We finetune the generator on a new instruction task: the task of predicting whether the answer is correct or wrong, given the question and document set. We first compute the log-probability associated with predicting “correct” or “wrong”, using the RAG-generator.
\[ y \leftarrow P_{\phi}(a, q, (d_{1},..., d_{M}))\]

We then minimize the cross-entropy loss based on the ground-truth label and the model’s assessment of the answer’s veracity:
\begin{equation}
loss(\phi) = C.E(y, \hat{y})
\label{eqn:calibrationloss}
\end{equation}

\bibliography{custom}
\bibliographystyle{acl_natbib}

\appendix

\section{Appendix I}
\label{sec:appendix}
\textbf{Prompt Template for Extractive/YN Formats}
“Given the passage, generate an extractive question answer pair, relevant to the passage. The answer to the question should be a span from the given passage or a yes/no answer. Provide a rationale for the answer as well.
<Exemplars>  \\
Passage: [X]”. 
\\
\\
\textbf{Prompt Template for Abstractive Formats}
“Given the passage, generate an abstractive question answer pair, relevant to the passage. The answer should be grounded in the passage.
<Exemplars> \\
Passage: [X]”.

\end{document}